# Uncertainty Propagation in Deep Neural Network Using Active Subspace

Weiqi Ji [a], Zhuyin Ren [a] and Chung K. Law [a,b]

[a] Tsinghua University, Beijing, China [b] Princeton University, NJ, USA

weiqiji@mit.edu (W. Ji), zhuyinren@tsinghua.edu.cn (Z. Ren),

cklaw@princeton.edu (C. K. Law)

## Abstract

The inputs of deep neural network (DNN) from real-world data usually come with uncertainties. Yet, it is challenging to propagate the uncertainty in the input features to the DNN predictions at a low computational cost. This work employs a gradient-based subspace method and response surface technique to accelerate the uncertainty propagation in DNN. Specifically, the active subspace method is employed to identify the most important subspace in the input features using the gradient of the DNN output to the inputs. Then the response surface within that low-dimensional subspace can be efficiently built, and the uncertainty of the prediction can be acquired by evaluating the computationally cheap response surface instead of the DNN models. In addition, the subspace can help explain the adversarial examples. The approach is demonstrated in MNIST datasets with a convolutional neural network. Code is available at: https://github.com/jiweiqi/nnsubspace.

## 1. Introduction

Deep neural networks (DNNs) have demonstrated impressive performance over the years in many fields of research. The applications include object classification (Krizhevsky, Sutskever and Hinton 2012; He et al. 2016), semantic segmentation (Long, Shelhamer and Darrell 2015), activity recognition/detection (Tran et al. 2015) and speech recognition (Hinton et al. 2012) to name a few. Despite these successes, DNNs prove to be not very robust to noise in the input (Tang and Eliasmith 2010). Recent works on adversarial perturbations (Szegedy et al. 2013) clearly show that small imperceptible perturbations to the input can result in a drastic negative effect on classification performance. Real world data come with uncertainties and quantifying the effect of input noise to DNN predictions becomes vital for those safety-critical systems such as autonomous driving (Gal and Ghahramani 2016). On the other hand, training networks with random noise applied to their inputs can enhance the robustness of networks (Sietsma and Dow 1991). Therefore, understanding the response of the network output to the input noise is also useful for training the network, and provides insight on searching adversarial examples (Goodfellow, Shlens and Szegedy 2014).

Thus, in this work, we introduce the active subspace method (Constantine, Dow and Wang 2014) to identify the low-dimensional subspace from the high-dimensional input features in DNN. Along the active subspace, the perturbation of the inputs substantially changes the output while the perturbation in the complementary subspace has little effect on the output. The active subspace can be identified via the gradient of DNN output w.r.t. the inputs, which can be evaluated efficiently via backpropagation. We shall show that one or two-dimensional active subspace can reasonably capture most of the variations of the output in the MNIST dataset under a moderate level of perturbation. Then the uncertainty of the output can be efficiently estimated with few samples through building low-dimensional response surface against the active subspace.

## 2. Related Work

Monte Carlo (MC) sampling is the most straightforward approach that can be used to calculate the uncertainty of the output. It is based on randomly drawing a number of samples from the inputs distribution and then feed those samples to the network. The probabilistic density function, as well as the first and second moment of the output, can be estimated through those predictions. However, MC requires a large number of samples due to its slow convergence rate.

Several approaches have been proposed to estimate the uncertainty of output efficiently, and they can be divided into two categories. One is layer-wise uncertainty propagation (Bibi, Alfadly and Ghanem 2018; Astudillo and Neto 2011) and the other one is using Unscented Transform (Simon and Uhlmann 1997; Abdelaziz et al. 2015). In the layer-wise approach, the first and second moment of a single layer's output is analytically expressed as the layer inputs uncertainty under certain assumptions and then the uncertainty of the input is propagated to the final output layer-by-layer. However, the performance degrades when the network is deep as the error accumulates through layers. The Unscented Transform is based on the Unscented Kalman Filter (Simon and Uhlmann 1997), which assumes that the output follows a Gaussian distribution and its first and second moment can be accurately estimated via a series of weighted deterministic samples. However, the Unscented Transform requires $2d + 1$ samples where $d$ is the dimension of the input features. The computational cost could be significant for DNN with high-dimensional input features.

## 3. Approach

We denote the input vector as $\mathbf{x}$ following the probabilistic distribution of $\pi_{\mathbf{x}}$. The deterministic DNN maps the inputs to output and is denoted as $f(\mathbf{x})$. The active subspace (AS) methodology is detailed in (Constantine, Dow and Wang 2014) with algorithms, a rigorous error analysis, and demonstrations. It has been applied to various engineering field in the context of forward uncertainty propagation (Constantine et al. 2015; Ji et al. 2018) and Bayesian inference (Constantine, Kent and Bui-Thanh 2016). Here we simply state the basic concepts. The AS method seeks an $r$-dimensional subspace in the $d$-dimensional input feature space that describes most of the variation of $f(\mathbf{x})$. The idea is to find a low-dimensional approximation of $f(\mathbf{x})$ as

$$f(\mathbf{x}) \approx g(\mathbf{x}_r), \mathbf{x}_r = \boldsymbol{S}^T \mathbf{x}, \tag{1}$$

where $g$ is a function of the $r$-dimensional input $\mathbf{x}_r$ with $r < d$, and $\boldsymbol{S}$ is an orthogonal matrix of size $d \times r$. The active subspace is defined as *span*$\{\boldsymbol{S}\}$. One way to identify the active subspace is to perform an eigenvalue decomposition of the matrix $\boldsymbol{C}$, defined as the expectation of the outer product of the gradient $\nabla f$ with itself, *i.e.*,

$$\boldsymbol{C} = \int \nabla f(\mathbf{x}) \nabla f(\mathbf{x})^T \pi_{\mathrm{x}}(\mathbf{x}) \mathrm{d}\mathbf{x} = \boldsymbol{W} \boldsymbol{\Lambda} \boldsymbol{W}^{\boldsymbol{T}}. \tag{2}$$

Note that $\boldsymbol{C}$ is symmetric, positive semi-definite, and of size $d \times d$. The unitary matrix $\boldsymbol{W}$ consists of the $d$ eigenvectors $\mathbf{w}_1, \ldots, \mathbf{w}_d$ and $\boldsymbol{\Lambda}$ is a diagonal matrix whose components are the eigenvalues $\lambda_1, \ldots \lambda_d$, sorted in descending order. If there is a gap in the eigenvalues, meaning $\lambda_r \gg \lambda_{r+1}$, then the function $f$ varies mostly along the first $r$ eigenvectors and is almost constant along the rest of the eigenvectors. The first $r$ eigenvectors are selected as *active* directions, *i.e.*, $\boldsymbol{S} \equiv [\mathbf{w}_1, \ldots, \mathbf{w}_r]$ ; and its complement $span\{\mathbf{w}_{\boldsymbol{r+1}}, \ldots, \mathbf{w}_d\}$ is identified as the *inactive* subspace. Then one can build a response surface, $RS(\mathbf{x}_r)$, with $\mathbf{x}_r$ as input, and it is chosen as the function $g$, *i.e.*,

$$f(\mathbf{x}) \approx g(\mathbf{x}_r) = RS(\mathbf{x}_r). \tag{3}$$

The matrix $\boldsymbol{C}$ can be approximated by MC simulations. The number of gradient evaluations of the forward model, $M$, variable increases logarithmically with $d$, i.e.,

$$M = \alpha\beta \log(d). \tag{4}$$

The constant $\alpha$ is the over-sampling factor and is recommended to be between 2 and 10, and $\beta$ should be larger than $r + 1$.

Once the active subspace is identified, various response techniques, such as Polynomial Fitting and Polynomial Chaos Expansion (PCE) (Conrad and Marzouk 2013) can be readily applied to the low-dimensional active subspace. The entire workflow for propagating the input uncertainty to the DNN output will be:

(**I**) Estimate the active subspace based on a small number, $M$, of samples drawn from $\pi_{\mathrm{x}}$, and evaluate the gradient of $f(\mathbf{x})$ for each sample.

(**II**) Build the response surface $RS(\mathbf{x}_r)$ to the low-dimensional variable $\mathbf{x}_r$.

(**III**) Estimate the distribution of the DNN output by evaluating a large number of samples drawn from $\pi_{\mathrm{x}}$ with $f(\mathbf{x})$ being approximated by the cheap response surface $RS(\mathbf{x}_r)$.

## 4. Experimental Result

In this section, we apply the active subspace method to the MNIST handwritten digital data (LeCun 1998). A (convolutional neural network) CNN, similar to LeNet (LeCun et al. 1998), with four hidden layers was trained on 60000 images to the accuracy of 99.28%, and the test accuracy of 98.86% on another 10000 test images. Dropout is applied for the regularization and the activation function of softplus instead of ReLU is implemented such that the DNN output is mathematically differentiable w.r.t the input features. The uncertainty of the image follows Gaussian distribution, which is centered at the original image with a constant variance for all of the pixel values. The uncertainty of each pixel value is independent with each other. For instance, the variance of the input feature vector $\mathbf{x}$ can be written as $\sigma^2 I_{784}$, in which $\mathbf{x}$ corresponds to the flatten vector of the image matrix, and $I$ is the identity matrix of size $784 \times 784$. We denoted the additive Gaussian noise as $\boldsymbol{\xi}$, then we have

$$\mathbf{x} = \mathbf{x}_0 + \sigma\boldsymbol{\xi} \tag{3}$$

in which $\boldsymbol{\xi}$ follows a standard Gaussian distribution, i.e., $\boldsymbol{\xi} \sim N\,(\mathbf{0}, I_{784})$. Noting that the distribution of the added noise is truncated to ensure the pixel value to be within [0, 255]. The output, corresponds to the predicted label of the original picture, is specified as the quantity of interest, although the predicted label might change after adding noise to the original picture. For example, a picture with the ground truth label of '6' is classified as '8', then the output corresponds to the category of '8', intead of '6', will be specified as the quantity of interest. The constants of $\alpha$ and $\beta$ in Eq. (4) are both specified as 10, and the number of evaluations of the gradient $f(\mathbf{x})$ will be $667 = 10 * 10 * \ln(28 * 28)$ for each image.

Figure 1 presents the eigenvalues and the summary plot of the output against the first active variable for an image from the test dataset, shown in Fig. 2. The first active variable defined as the projected value of noise onto the first active direction, i.e., $\mathbf{w}_1^T\boldsymbol{\xi}$. The spectrum suggests a one-dimensional active subspace for the cases of σ = 20 and 30, and a two-dimensional active subspace for the case of larger uncertainty σ = 50. The summary plot shows that the first active variable can capture the changes of the output with the noise input very well, although the scattering in the case of σ = 50 becomes significant. In all three cases, the map from input features to output is close to linear with weak nonlinear behavior, and a response surface with second order polynomial fitting is sufficient for those cases. Finally, the uncertainty of the output is acquired by evaluating 50000 samples via the response surface. The histogram suggest that the distribution of the output is close to Gaussian distribution.

As the DNN output changes most along the active direction, one can make adversarial examples by specifying the noise along the active direction. An example is shown in Fig.

2. The score of the output is significantly changed, although there is no visual difference between the original and the perturbated image.

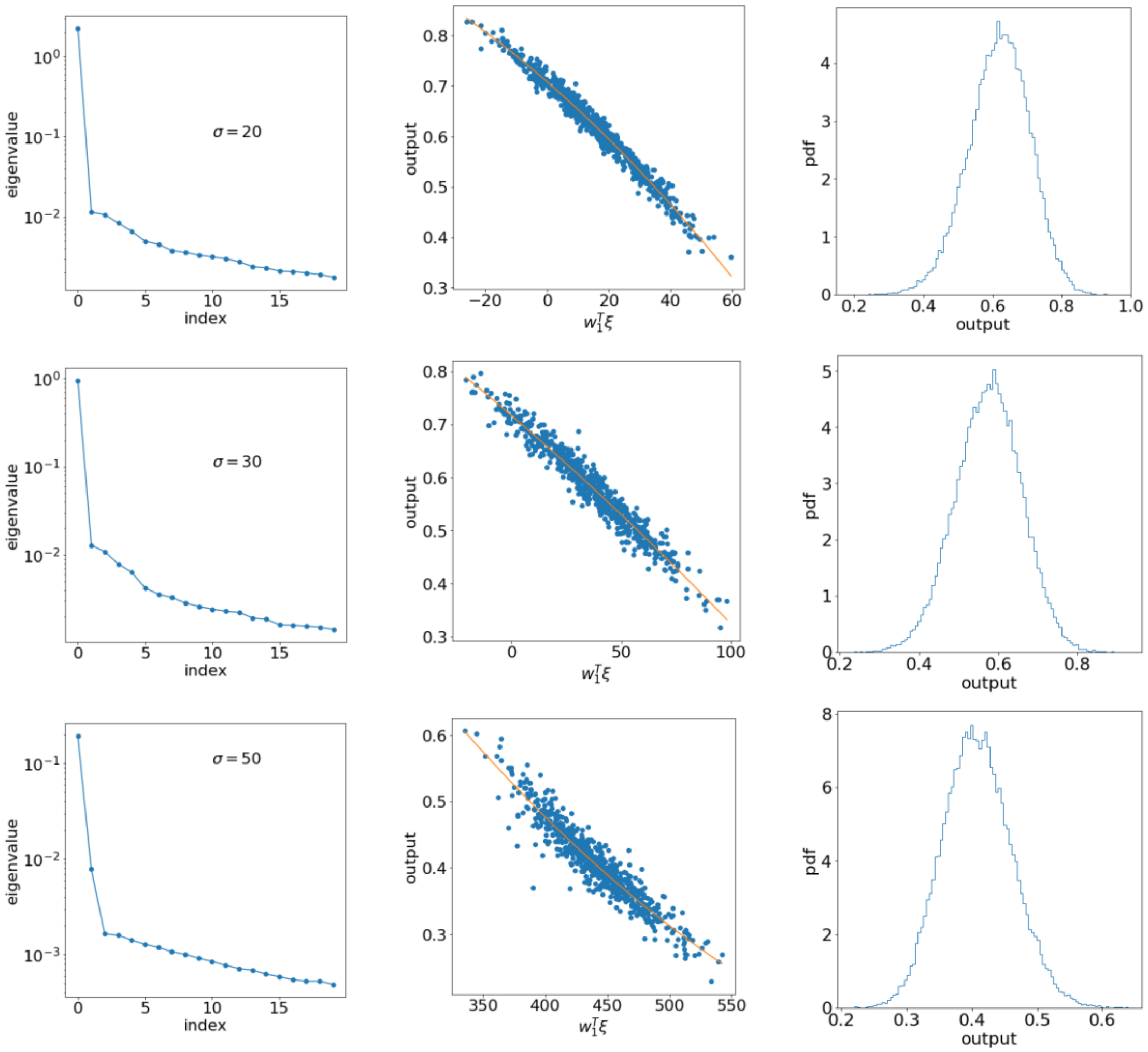

Figure 1. The spectrum of the eigenvalues, the summary plot against the first active variable along with the second order polynomial fitting curve, and the histogram of the network output based on the response surface.

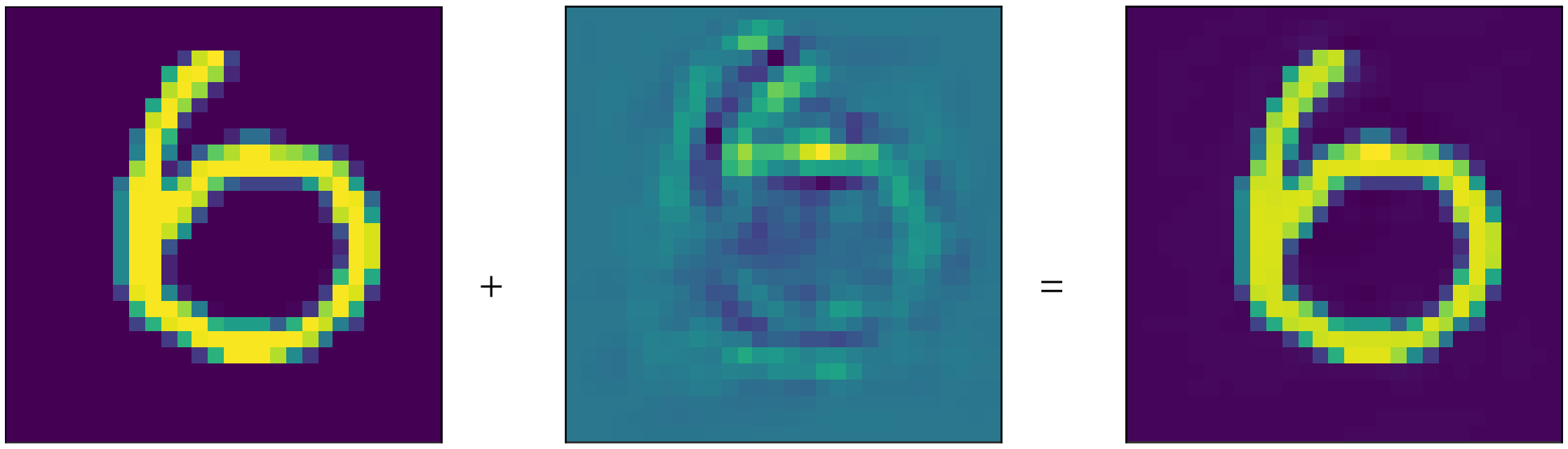

$\mathbf{x}$ {‘6’: 0.69 confidence}    $\boldsymbol{w}_1^T\boldsymbol{\xi} = 63.75$ {‘1’: 0.19 confidence }    $\mathbf{x}+\boldsymbol{w}_1^T\boldsymbol{\xi}$ {‘0’: 0.89 confidence }

Figure 2. The original image (left), addictive noise along the first active direction, and the image added with the noise. $\mathbf{w}_1^T$ corresponds to the active subspace for the case of σ = 20.

Finally, the accuracy of the computed output uncertainty through one-dimensional active subspace and second-order polynomial fitting is validated against MC sampling. Figure 3 shows the conditional plot of the standard deviations of the DNN output against its mean from 1000 test images. For each image, we use 50000 samples for the Monte Carlo integration, and the uncertainty of the input is specified as σ = 20. As expected, the uncertainty of the output is very small when the highest score of the prediction is close to 100%. Therefore, we shall focus on the images whose highest score is less than 0.9. Figure 4 presents the results from active subspace with response surface and the direct MC. In all three cases, the mean and the variance of the DNN output agree with MC very well. Assuming that the computational cost of the gradient evaluation is the same as the forward prediction of DNN and ignoring the cost of building response surface, the total cost of the current approach is lower than that of direct MC by two magnitudes.

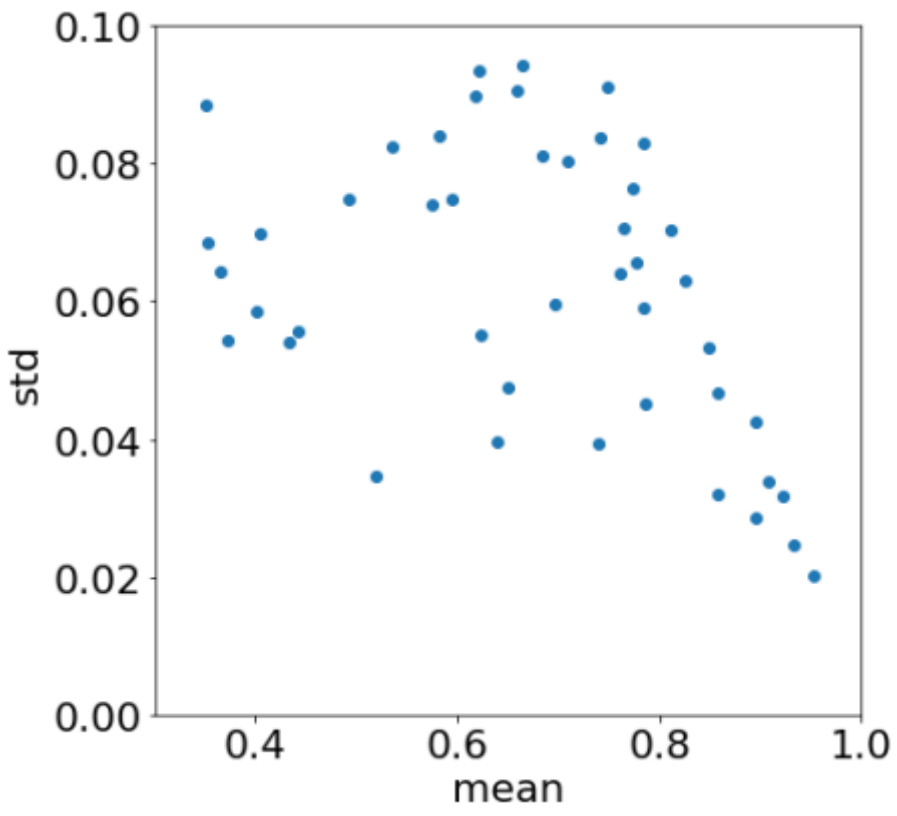

Figure 3. The conditional plot of the standard deviations of the output against its mean. The statistics of the output is acquired via direct Monte Carlo sampling with 50000 samples for each image under σ = 20.

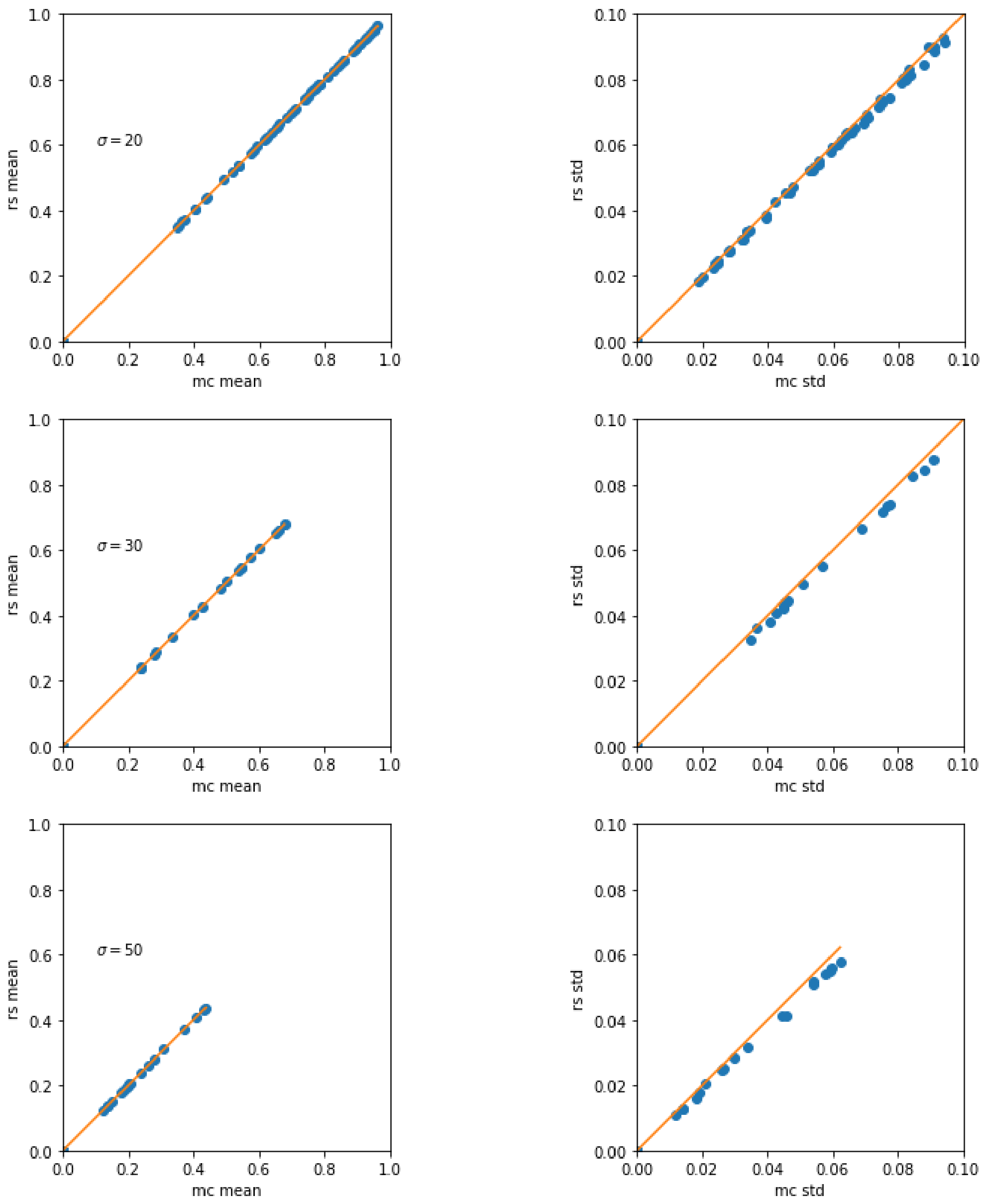

Figure 4. The mean and standard deviation of the output acquired via response surface (rs mean & std) against the ones from direct Monte Carlo sampling (mc mean & std).

## 5. Discussion

The current work shows that there exist one-dimensional active subspace from the map of the network inputs to the output for most of the test images under a moderate level of uncertainty, although some images show two-dimensional active subspace. Furthermore, the map between the output and inputs is close to linear for most of the images especially when the variance of the input uncertainty is small. These observations justify the argument made by (Goodfellow, Shlens and Szegedy 2014), that the linear nature of DNN w.r.t the perturbations to inputs can explain the adversarial examples rather than the nonlinear response. Regarding the cases where the active subspace is higher than one-dimension or the response of the output is nonlinear, the summary plot against the one and two-dimensional active subspace enable us to visualize the response of the output to the perturbations.

The number of gradient evaluations specified in the current work is very conservative, and it can be further reduced to facilitate the efficiency requirement of real-time prediction, such as the perception and end-to-end control in autonomous driving.

As a byproduct, the active subspace also reveals the global sensitivity of the output to the inputs (Constantine and Diaz 2017), *i.e.*, large components of the active direction corresponds to the influential features. This opens the possibility of developing subspace based metrics for attributing the prediction of DNN to its input features (Sundararajan, Taly and Yan 2017).

In the future, we shall apply the approach to more complex datasets, such as ImageNet (Deng et al. 2009), with much deeper networks, such as ResNet (He et al. 2016) and DenseNet (Huang et al. 2017). As the number of evaluations of gradients increases logarithmically with the dimension of inputs, the active subspace can also be applied to the uncertainty in the model parameters with hundreds of gradient evaluation. For example, ResNet50 contains about $2.5 \times 10^7$ model parameters, the number of gradient evaluation with both $\alpha$ and $\beta$ set as 5 will be 425. Although Monte Carlo dropout (Gal and Ghahramani 2016) is widely adopted for propagating the uncertainties in the network parameters to the network output, it assumes that the uncertainty of the model parameters follows a Bernoulli distribution. The approach presented in the current work is suitable for more complex distributions in the model parameters.

## 6. Conclusion

In this work, the active subspace is applied to identify the low-dimensional subspace in the map from the input features to the network output. With backpropagation, the active subspace can be identified with hundreds of gradient evaluations. One dimensional active subspace and linear response are observed for most of the test images under a moderate level of perturbations to the inputs, although two-dimensional active subspace and non-linear response are also observed for large input uncertainties. Those findings are useful for the explanations and searching of adversarial examples. In conjunction with response surface methodology, the statistics of the DNN output can also be efficiently acquired, and the computational cost is less than the direct Monte Carlo method by two magnitudes in the MNIST dataset.

## 7. Acknowledgment

This work was supported by the National Natural Science Foundation of China No. 91441202.